\documentclass[10pt,twocolumn,letterpaper]{article}

\usepackage{wacv}
\usepackage{times}
\usepackage{epsfig}
\usepackage{graphicx}
\usepackage{amsmath}
\usepackage{amssymb}

\def\wacvPaperID{9} 

\wacvfinalcopy 

\ifwacvfinal
\def\assignedStartPage{1} 
\fi

\ifwacvfinal
\usepackage[breaklinks=true,bookmarks=false]{hyperref}
\else
\usepackage[pagebackref=true,breaklinks=true,colorlinks,bookmarks=false]{hyperref}
\fi

\ifwacvfinal
\else
\fi

\begin{document}

\title{Spatiotemporal Motion Synchronization for Snowboard Big Air}

\author{Seiji Matsumura, Naoki Saijo, Makio Kashino\\
NTT Communication Science Laboratories\\
3-1, Morinosato Wakamiya, Atsugi-shi, Kanagawa, Japan\\
{\tt\small \{seiji.matsumura.yh, naoki.saijo, makio.kashino.ft\}@hco.ntt.co.jp}
\and
Dan Mikami\\
Department of Computer Science, Kogakuin University\\
1-24-2, Nishi-Shinjuku, Shinjuku-ku, Tokyo, Japan\\
{\tt\small mikami.dan@cc.kogakuin.ac.jp}
}

\maketitle

\begin{abstract}
    During the training for snowboard big air, one of the most popular winter sports, athletes and coaches extensively shoot and check their jump attempts using a single camera or smartphone. However, by watching videos sequentially, it is difficult to compare the precise difference in performance between two trials. Therefore, side-by-side display or overlay of two videos may be helpful for training. To accomplish this, the spatial and temporal alignment of multiple performances must be ensured. In this study, we propose a conventional but plausible solution using the existing image processing techniques for snowboard big air training. We conducted interviews with expert snowboarders who stated that the spatiotemporally aligned videos enabled them to precisely identify slight differences in body movements. The results suggest that the proposed method can be used during the training of snowboard big air.
\end{abstract}

\section{Introduction}

Sports performance involves complex body movements. Athletes observe their own movements as well as those of others to improve their skill and performance. In snowboard big air, specifically, athletes perform tricks that require highly skilled body movements in the air using an artiﬁcial snow jump stand. In the big air competition, both the skill and psychophysiological state of an athlete affect their performance~\cite{10.3389/fspor.2021.712439}; therefore, comparing the movements during training and the competition can prove to be helpful in improving the skills of athletes.

In big air training, it is common for athletes to record their own jumps with a single camera or smartphone for training and enhancing their skills. However, as the athletes check these videos manually, the usefulness of such data is limited. In contrast, in other sports, videos are used efficiently; for example, in baseball, a system called Statcast~\cite{8074554} is used to measure the speed and rotation of the ball thrown by the pitcher, the speed and angle of the ball hit by the batter, and the position of player. The Tracab system~\cite{10.3389/fspor.2021.624475} is widely used in soccer to obtain information regarding the position of the ball and players. While these measured data are mainly used for strategic planning in a team, their use by individual players to improve their own skills is limited. In addition, it is difficult to implement these systems in snowboarding events that are held in outdoor winter environments because the measurement devices of the systems must be ﬁxed in a certain location.

To improve skills in a more intuitive way, Chua et al. proposed a system that measures the movements of TaiChi practitioners via motion capture and presents the model movements by matching the movements of a practitioner in real time on the head-mounted display worn by them during training~\cite{1191125}. During golf training, Ikeda et al. perceived the player movements; subsequently, they presented the model movements as well as the player movements as a virtual shadow in real time so that the player can easily understand the difference in movements~\cite{8798196}.

However, these approaches involve two major limitations in their application to snowboard big air. The first limitation is measuring the motions of an athlete; because this event is held in a vast snowy outdoor environment, it is difficult to sense the posture of an athlete. Even though posture estimation based on deep learning is now becoming popular~\cite{10.3389/fspor.2020.00050}, it is still difficult because snowboard clothing does not adhere tightly to the body. The second limitation is the characteristics of the motions of an athlete in big air; athletes move quickly and cover larger distances in a few seconds with complex postural changes to perform tricks in the air. Therefore, it is impractical to provide real-time visual feedback to snowboarders.

As another example of a more intuitive approach to improving skills, Mikami et al. proposed a system that shows the movements of a practitioner in temporal synchronization with professionally modeled movements immediately after the practitioner's actual movements~\cite{Mikami2013AVF}. The system synchronizes the practitioner's movements with those of the model based on a motion feature called the motion history image and presents them immediately after the movements. If the difference between the practice and model movements is relatively large, these movements are synchronized based on rough motion features without posture estimation. This approach can be advantageous for big air training because athletes can quickly compare multiple jump attempts and intuitively capture the differences in performance.

Because this method is assumed to be used for analyzing movements performed at an exact location using a camera fixed at a specific position, it cannot be easily implemented in snowboard big air, which is held outdoors; moreover, it is difﬁcult to keep the camera in the same location. In this method, the difference in viewpoints in the videos cannot be controlled; therefore, it is difficult to compare two jump performances in two different videos intuitively.

In this study, we propose a novel method in which the videos of different jump attempts were temporally synchronized by aligning the viewpoints; subsequently, they were shown side-by-side or overlaid. After implementing this method, we interviewed expert snowboarders and conﬁrmed that the effect of spatiotemporal synchronization on training is promising.

The remainder of this paper is organized as follows. We propose spatiotemporal synchronization to help train athletes to detect detailed differences in motions as the first step. Section 2 provides an overview of the proposed system. Section 3 presents the current implementation. Section 4 presents the results of spatiotemporal synchronization and interviews with expert snowboarders. Finally, Section 5 presents the conclusions of our study.

\section{System overview}
\label{sec:system_overview}

This system takes two trial videos as input and generates a video for comparison, which makes it easy for athletes to understand the difference between movements. This system can mainly be used to create a video for comparing the videos of the model and athletes. Because snowboard big air is often performed outdoors during the winter, it is difficult to record both input trial videos from the same viewpoint. Therefore, the proposed method has two main functions, namely, temporal and spatial synchronization. The system then outputs a video that simultaneously presents two spatiotemporally synchronized trials (see Figure 1).

In temporal synchronization, the base frame for synchronization in both input videos is detected and the offset value is determined such that the frames occur simultaneously. Consequently, the trial timing between the two videos is aligned.

Spatial synchronization is a viewpoint correction method. The system uses one of the videos as a reference and corrects the viewpoint of the other video with respect to that of the reference video. This process enables the comparison between the two videos from the same viewpoint.

\begin{figure}[t]
\begin{center}
\includegraphics[width=1.0\linewidth]{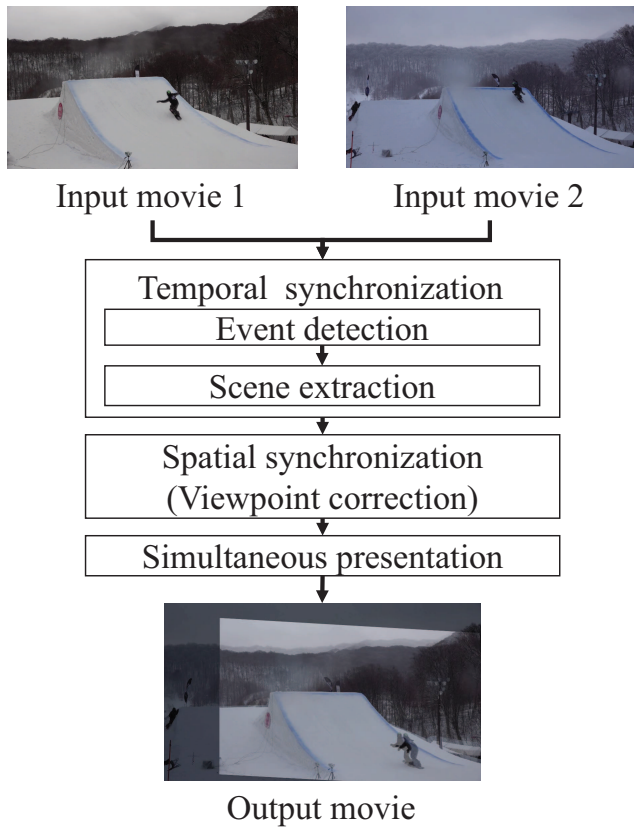}
\end{center}
   \caption{Flowchart of the proposed system.}
\label{fig:long}
\label{fig:onecol}
\end{figure}

\section{Implementation}
\label{sec:implementation}

All videos used for input were captured using Sony FDR-AX700. The resolution of the videos was 1920 × 1080 pixels with a frame rate of 120 fps.

Now, we describe the implementation of temporal and spatial synchronization. In the temporal synchronization process, the system first automatically detects an event using background subtraction. The system sets up a linear region at the top of the jump stand, as shown by the red line in Figure 2. It calculates the luminance difference between the reference and input images in this region and selects the frame in which luminance difference is observed. In this study, the first frame was selected as the base point of the event. For the reference image, an image that did not include the snowboarder was used (see the left panel of Figure 2). Then, a video of a certain time width before and after the base frame was extracted. In this implementation, because the luminance difference was obtained at the top of the jump stand, the take-off moment became the base point, and a certain time before and after the take-off became the extracted section.

\begin{figure}[t]
\begin{center}
\includegraphics[width=1.0\linewidth]{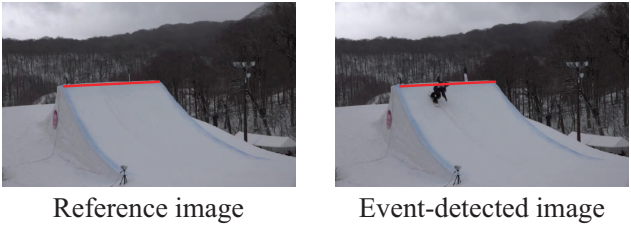}
\end{center}
   \caption{Automatic event detection using background subtraction.}
\label{fig:long}
\label{fig:onecol}
\end{figure}

For the spatial synchronization process, the system uses homography transformation to correct the viewpoint of one video to match that of the other. Specifically, as shown in Figure 3, we manually obtained the coordinates of the four corners of the jump stand in both videos. Using these coordinates, the system calculates a transformation matrix $H'$ based on the following equation and performs homography transformation. In the following equation, $(x, y)$ and $(x',y')$ denote the coordinates in the source and destination videos, respectively.

\begin{displaymath}
\begin{bmatrix}
X'\\
Y'\\
W'\\
\end{bmatrix}
= H'
\begin{bmatrix}
x\\
y\\
1\\
\end{bmatrix}
= 
\begin{bmatrix}
h'_{00} & h'_{01} & h'_{02}\\
h'_{10} & h'_{11} & h'_{12}\\
h'_{20} & h'_{21} & h'_{22}\\
\end{bmatrix}
\begin{bmatrix}
x\\
y\\
1\\
\end{bmatrix}
\end{displaymath}

\begin{displaymath}
\begin{bmatrix}
x'\\
y'\\
\end{bmatrix}
=
\begin{bmatrix}
X'/W'\\
Y'/W'\\
\end{bmatrix}
\end{displaymath}

However, homography transformation may involve spatial distortion, and the distortion in the posture of athletes owing to this transformation can lead to discomfort in the field.

\begin{figure}[t]
\begin{center}
\includegraphics[width=1.0\linewidth]{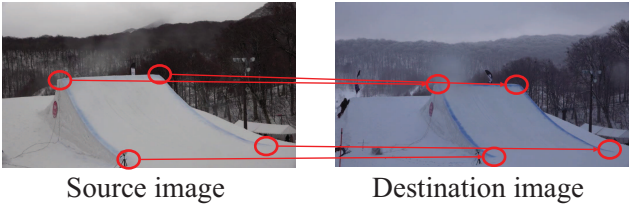}
\end{center}
   \caption{Viewpoint correction by homography transformation.}
\label{fig:long}
\label{fig:onecol}
\end{figure}

Finally, two spatiotemporally synchronized trials are presented simultaneously in either side-by-side or superimposed form.

\section{Results and Discussion}
\label{sec:resultanddiscussion}

Figure 4 shows the output video obtained using the proposed method. Two videos synchronized in time and space were presented side by side or superimposed simultaneously. Table 1 summarizes the results of quantitative evaluation of the time synchronization accuracy. To evaluate the accuracy, we calculated the difference in errors in the base frame detected in each input video; the smaller the difference, the more temporally synchronized the detected frames. For comparison, we implemented a method for detecting the moment when a body passes the top of a jump stand using PoseNet~\cite{kendall2015posenet}, a widely used pose estimation method for event detection. The results showed that the synchronization accuracy of event detection by background subtraction adopted in the proposed method was high.

\begin{figure}[t]
\begin{center}
\includegraphics[width=1.0\linewidth]{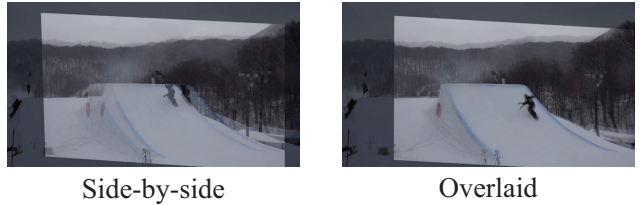}
\end{center}
   \caption{Examples of output videos.}
\label{fig:long}
\label{fig:onecol}
\end{figure}

\begin{table}
\begin{center}
\begin{tabular}{|l|c|}
\hline
Method & Difference in errors [frames]\\
\hline\hline
Background subtraction & 1.3 ± 1.7 \\
Pose estimation & 6.8 ± 9.1 \\
\hline
\end{tabular}
\end{center}
\caption{Difference in the error of the base frame. Mean ± SD for all 10 trials.}
\end{table}

We interviewed four expert snowboarders to conﬁrm whether the videos created by the proposed method would be helpful in snowboard training. Three of them were professional snowboarders, and the other was a semi-professional snowboarder. In the interviews, we ﬁrst instructed the participants to watch the two videos in sequence as if they were using the videos in daily training; then, we asked them about the difference in performance of their tricks. Subsequently, we asked the participants to watch the videos created using the proposed method and discuss if they had any new insights compared to the conventional method and whether the proposed method would be helpful in training.

For the ﬁrst interview topic, we received responses stating that by watching the two videos in sequence, the snowboarders could detect relatively significant differences in the jumping movements such as the jump distance, quality of rotation axis, tendency to load the board when taking off, and upper body tilt during the approach phase. This may be because they usually watch videos in sequence during training. In contrast, the side-by-side video generated using the proposed method enabled them to recognize these differences more clearly, and they answered that it would be easier for beginners to identify the quality of their jumps. Moreover, they stated that the superimposed video could help them in recognizing detailed spatiotemporal differences in the movements, such as the time of initiating the preceding movement in the approach phase and the difference in rotational angle of the arm. This suggests that athletes can notice minimal but important differences in body movements during performance using our method.

Interestingly, some of the interviewees said that the difference in actions appeared smaller in the proposed method than that in general usage. Although the reason for this remains unclear, there may be at least two possible explanations. First, the motion difference is overestimated when it is recognized from videos shown sequentially. Another possibility is that irrelevant information from the body motions in the side-by-side or superimposed images  makes the difference in movements relatively small. Further studies on the cognitive aspects of detecting differences in body movements from multiple videos are required in the future.

Contrary to expectations, no one noticed spatial postural distortions induced by homography transformation, suggesting that the system could be practically used if the difference in camera angle was kept within the range of our sample videos. Thus, the available range of difference in angles should be estimated in future studies to ensure that the proposed method can be applied to actual training scenarios in snowboard big air.

For the second interview topic, all interviewees stated that the videos produced by the proposed method would be helpful in training. Additionally, they proposed that it would be useful in two specific situations. First, it would be useful to compare the jumps observed in practice to those in competitions. In their experience, they could not deliver the performance demonstrated during practice in the competitions, and the proposed method can be used to verify the performance difference between the practice and competition. Second, this method can be used to learn new tricks. When they try a new trick, they often imagine the movement in detail before attempting it. Therefore, the proposed method would be useful for them to compare the video of the completed trick with that of their practice and determine a correct image of the movement.

The proposed method can be useful for big air athletes. Finally, we present two future research directions. First, the synchronization process can be automated. In this study, the corner points for homography transformation were detected manually in the spatial synchronization process. For the proposed method to be widely used, it is desirable to automate the entire process. Second, support for hand-held shooting should be considered. In this study, we used videos captured by a fixed-point camera placed on snow. However, in actual training, hand-held shooting with a smartphone is more popular. Moreover, it is necessary to consider the effects of camera shaking to cope with hand-held shooting.

Although there are limitations and future research is required, the proposed spatiotemporal motion synchronization method is expected to be more effective for snowboard big air training.

\section{Conclusion}
\label{conclusion}

In this study, we proposed a spatiotemporal motion synchronization method for snowboard big air and interviewed expert snowboarders regarding their experience with its usefulness. The results of the interviews suggested that the proposed method can help in training athletes to detect detailed spatial and temporal differences in motions.

{\small
\bibliographystyle{ieee_fullname}
\bibliography{egpaper}
}

\end{document}